\documentclass{bmvc2k}
\usepackage{graphicx}
\usepackage{caption}
\usepackage{units}

\title{Real-Time Intensity-Image Reconstruction for Event Cameras Using Manifold Regularisation}
\addauthor{Christian Reinbacher}{reinbacher@icg.tugraz.at}{1}
\addauthor{Gottfried Graber}{graber@icg.tugraz.at}{1}
\addauthor{Thomas Pock$^{\mbox{\sffamily\scriptsize 1,}}$}{pock@icg.tugraz.at}{2}

\addinstitution{
 Graz University of Technology\\
  Institute for Computer Graphics\\and Vision
}
\addinstitution{
 Austrian Institute Of Technology\\
 Vienna
}

\def\eg{\emph{e.g}\bmvaOneDot}

\def\etal{\emph{et al}\bmvaOneDot}
\def\ie{\emph{i.e}\bmvaOneDot}

\graphicspath{{images/}{images/compare_manifold/}{images/compare_bardow/}{images/compare_dslr/}}
\usepackage{amsmath}
\usepackage{amssymb}
\usepackage{mathtools}
\newcommand{\Fig}{Fig.}

\newcommand{\refFig}[1]{\mbox{\Fig~\ref{#1}}}

\newcommand{\refFigsee}[1]{see \refFig{#1}}

\newcommand{\Eqn}{Eqn.}

\newcommand{\refEqn}[1]{\mbox{\Eqn~(\ref{#1})}}

\newcommand{\Sec}{Section}

\newcommand{\refSec}[1]{\Sec~\ref{#1}}

\usepackage[textwidth=2cm,figwidth=\columnwidth,colorinlistoftodos]{todonotes}
\makeatletter
\renewcommand{\todo}[2][]{\tikzexternaldisable\@todo[#1]{#2}\tikzexternalenable}
\makeatother

\newcounter{mycomment} 

\DeclareMathOperator*{\argmin}{\arg\!\min}

\runninghead{Reinbacher \bmvaEtAl}{Real-Time Image Reconstruction for Event Cameras}
\begin{document}

\maketitle

\begin{abstract}
Event cameras or neuromorphic cameras mimic the human perception system as they
measure the per-pixel {\em intensity change} rather than the actual {\em
intensity level}. In contrast to traditional cameras, such cameras capture new
information about the scene at MHz frequency in the form of sparse events. The
high temporal resolution comes at the cost of losing the familiar per-pixel
intensity information. In this work we propose a variational model that
accurately models the behaviour of event cameras, enabling reconstruction of
intensity images with arbitrary frame rate in real-time. Our method is
formulated on a per-event-basis, where we explicitly incorporate information
about the asynchronous nature of events via an {\em event manifold} induced by
the relative timestamps of events. In our experiments we verify that solving the
variational model on the manifold produces high-quality images without
explicitly estimating optical flow.
\end{abstract}

\section{Introduction}
In contrast to standard CMOS digital cameras that operate on frame basis,
neuromorphic cameras such as the Dynamic Vision Sensor
(DVS)~\cite{Lichtsteiner2008} work asynchronously on a pixel level. Each pixel
measures the incoming light intensity and fires an {\em event} when the absolute
change in intensity is above a certain threshold (which is why those cameras are
also often referred to as {\em event cameras}). The time resolution is in the
order of $\mu s$. Due to the sparse nature of the events, the amount of data
that has to be transferred from the camera to the computer is very low, making
it an energy efficient alternative to standard CMOS cameras for the tracking of
very quick movement \cite{Delbruck2007,Wiesmann2012}. While it is appealing that
the megabytes per second of data produced by a digital camera can be compressed
to an asynchronous stream of events, these events can not be used directly in
computer vision algorithms that operate on a frame basis. In recent years, the
first algorithms have been proposed that transform the problem of camera pose
estimation to this new domain of time-continuous events
\eg~\cite{Benosman2014,Gallego2015,Kim2014,Mueggler2014,Mueggler2015,Weikersdorfer2013},
unleashing the full potential of the high temporal resolution and low latency of
event cameras. The main drawback of the proposed methods are specific
assumptions on the properties of the scene or the type of camera movement.

\begin{figure}[t!]
\begin{center}
  \subfigure[Raw Events]{\includegraphics[width=0.28\textwidth]{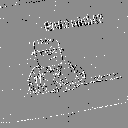}\label{subfig:events}}
  \subfigure[Reconstructed Image]{\includegraphics[width=0.28\textwidth]{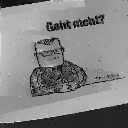}\label{subfig:image}}
  \subfigure[Event Manifold]{\includegraphics[width=0.28\textwidth]{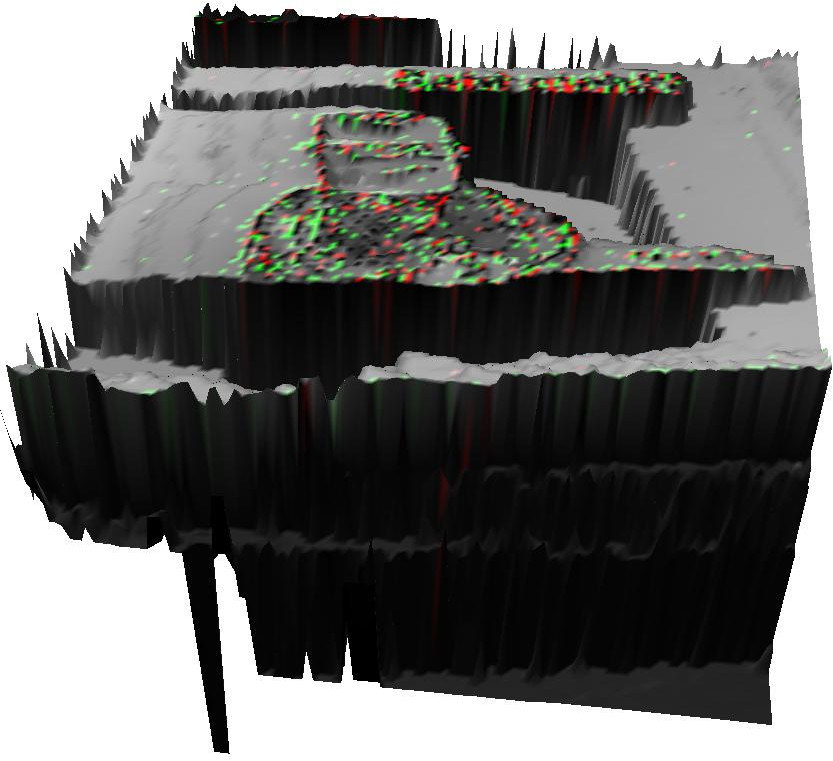}\label{subfig:manifold}}
	\caption{Sample results from our method. The image \subref{subfig:events} shows the raw events and \subref{subfig:image} is the
          result of our reconstruction. The time since the last event has happened
	for each pixel is depicted as a surface in \subref{subfig:manifold} with the positive and negative
	events shown in green and red respectively. }
	\label{fig:teaser}
\end{center}
\end{figure}

\paragraph*{Contribution}
In this work we aim to bridge the gap between the time-continuous domain of
events and frame-based computer vision algorithms. We propose a simple method
for intensity reconstruction for neuromorphic cameras (\refFigsee{fig:teaser}
for a sample output of our method). In contrast to very recent work on the same
topic by Bardow \etal~\cite{Bardow2016}, we formulate our algorithm on an
event-basis, avoiding the need to simultaneously estimate the optical flow. We
cast the intensity reconstruction problem as an energy minimisation, where we
model the camera noise in a data term based on the {\em generalised
Kullback-Leibler divergence}. The optimisation problem is defined on a manifold
induced by the timestamps of new events (\refFigsee{subfig:manifold}). We show
how to optimise this energy using variational methods and achieve real-time
performance by implementing the energy minimisation on a graphics processing
unit (GPU). We release software to provide live intensity image reconstruction
to all users of DVS
cameras\footnote{\url{https://github.com/VLOGroup/dvs-reconstruction}}. We
believe this will be a vital step towards a wider adoption of this kind of
cameras.

\section{Related Work}
Neuromorphic or event-based cameras receive increasing interest from the
computer vision community. The low latency compared to traditional cameras make
them particularly interesting for tracking rapid camera movement. Also more
classical low-level computer vision problems are transferred to this new domain
like optical flow estimation, or image reconstruction as proposed in this work.
In this literature overview we focus on very recent work that aims to solve
computer vision tasks using this new camera paradigm.
We begin our survey with a problem that benefits the most from the temporal
resolution of event cameras: camera pose tracking. Typical simultaneous
localisation and mapping (SLAM) methods need to perform image feature matching
to build a map of the environment and localise the camera within
\cite{Hartmann2013}. Having no image to extract features from means, that the
vast majority of visual SLAM algorithms can not be readily applied to
event-based data. Milford \etal~\cite{Milford2015} show that it is possible to
extract features from images that have been created by accumulating events over
time slices of \unit[1000]{ms} to perform large-scale mapping and localisation
with loop-closure. While this is the first system to utilise event cameras for
this challenging task, it trades temporal resolution for the creation of images
like \refFig{subfig:events} to reliably track camera movement.

A different line of research tries to formulate camera pose updates on an event
basis. Cook \etal~\cite{Cook2011} propose a biologically inspired network that
simultaneously estimates camera rotation, image gradients and intensity
information. An indoor application of a robot navigating in 2D using an event
camera that observes the ceiling has been proposed by Weikersdorfer
\etal~\cite{Weikersdorfer2013}. They simultaneously estimate a 2D map of events
and track the 2D position and orientation of the robot. Similarly, Kim
\etal~\cite{Kim2014} propose a method to simultaneously estimate the camera
rotation around a fixed point and a high-quality intensity image only from the
event stream. A particle filter is used to integrate the events and allow a
reconstruction of the image gradients, which can then be used to reconstruct an
intensity image by Poisson editing. All methods are limited to 3 DOF of camera
movement. A full camera tracking has been shown in
\cite{Mueggler2014,Mueggler2015} for rapid movement of an UAV with respect to a
known 2D target and in \cite{Gallego2015} for a known 3D map of the environment.

Benosman \etal~\cite{Benosman2014} tackle the problem of estimating optical flow
from an event stream. This work inspired our use of an event manifold to
formulate the intensity image reconstruction problem. They recover a motion
field by clustering events that are spatially and temporally close. The motion
field is found by locally fitting planes into the event manifold. In experiments
they show that flow estimation works especially well for low-textured scenes
with sharp edges, but still has problems for more natural looking scenes. Very
recently, the first methods for estimating intensity information from event
cameras without the need to recover the camera movement have been proposed.
Barua \etal~\cite{Barua2016} use a dictionary learning approach to map the
sparse, accumulated event information to infer image gradients. Those are then
used in a Poisson reconstruction to recover the log-intensities. Bardow
\etal~\cite{Bardow2016} proposed a method to simultaneously recover an intensity
image and dense optical flow from the event stream of a neuromorphic camera. The
method does not require to estimate the camera movement and scene
characteristics to reconstruct intensity images. In a variational energy
minimisation framework, they concurrently recover optical flow and image
intensities within a time window. They show that optical flow is necessary to
recover sharp image edges especially for fast movements in the image. In
contrast, in this work we show that intensities can also be recovered without
explicitly estimating the optical flow. This leads to a substantial reduction of
complexity: In our current implementation, we are able to reconstruct $>500$
frames per second. While the method is defined on a per-event-basis, we can
process blocks of events without loss in image quality. We are therefore able to
provide a true live-preview to users of a neuromorphic camera.

\section{Image Reconstruction from Sparse Events}
We have given a time sequence of events $(e^n)_{n=1}^N$ from a neuromorphic
camera, where $e^n=\{x^n,y^n,\theta^n,t^n\}$ is a single event consisting of the
pixel coordinates $(x^n,y^n)\in \Omega \subset \mathbb{R}^2$, the polarity
$\theta^n\in \{ -1,1\}$ and a monotonically increasing timestamp $t^n$.

A positive $\theta^n$ indicates that at the corresponding pixel the intensity
has increased by a certain threshold $\Delta^+>0$ in the log-intensity space.
Vice versa, a negative $\theta^n$ indicates a drop in intensity by a second
threshold $\Delta^->0$. Our aim is now to reconstruct an intensity image $u^n\ :
\Omega \to \mathbb{R}_+$ by integrating the intensity changes indicated by the
events over time.

Taking the $\exp(\cdot)$, the update in intensity space caused by one event
$e^n$ can be written as
\begin{equation}\label{eqn:intensity_update}
  f^{n}(x^n,y^n)= u^{n-1}(x^n,y^n) \cdot \begin{cases}
    c_1& \mbox{if } \theta^n>0\\
    c_2& \mbox{if } \theta^n<0
  \end{cases},
\end{equation}
where $c_1=\exp(\Delta^+)$, $c_2=\exp(-\Delta^-)$. Starting
from a known $u^0$ and assuming no noise, this integration procedure will
reconstruct a perfect image (up to the radiometric discretisation caused by
$\Delta^\pm$). However, since the events stem from real camera hardware, there
is noise in the events. Also the initial intensity image $u^0$ is
unknown and can not be reconstructed from events alone. Therefore the
reconstruction of $u^n$ from $f^n$ can not be solved without imposing some
regularity in the solution. We therefore formulate the
intensity image reconstruction problem as the solution of the optimisation
problem
\begin{equation}
\label{eqn:overall_optimization_problem}
u^n = \argmin_{u\in C^1(\Omega,\mathbb{R}_+)}\left[ E(u) = D(u,f^n) + R(u)\right]\ ,
\end{equation}
where $D(u,f^n)$ is a {\em data term} that models the camera noise and $R(u)$ is
a {\em regularisation term} that enforces some smoothness in the solution.
In the following section we will show how we can
utilise the timestamps of the events to define a manifold which guides a
variational model and detail our specific choices for data term and
regularisation.

\subsection{Variational Model on the Event Manifold}
\label{sec:manifold}
Moving edges in the image cause events once a change in logarithmic intensity is
bigger than a threshold. The collection of all events $(e^n)_{n=1}^N$ can be
recorded in a spatiotemporal volume $V\subset \Omega \times T$. $V$ is very
sparsely populated, which makes it infeasible to directly store it. To alleviate
this problem, Bardow et al.~\cite{Bardow2016} operate on events in a fixed time
window that is sliding along the time axis of $V$. They simultaneously optimise
for optical flow and intensities, which are tightly coupled in this volumetric
representation.

\paragraph*{Regularisation Term}
As in \cite{Benosman2014}, we observe that events lie on a lower-dimensional
manifold within $V$, defined by the most recent timestamp for each pixel
$(x,y)\in \Omega$. A visualisation of this manifold for a real-world
scene can be seen in \refFig{subfig:manifold}. Benosman \etal~\cite{Benosman2014}
fittingly call this manifold the {\em surface of active events}.
We propose to incorporate the surface of active events into our method by
formulating the optimisation \emph{directly on the manifold}.
Our intuition is, that parts of the scene that have no or little texture will
not produce as many events as highly textured areas. Regularising an image
reconstructed from the events should take into account the different ``time
history'' of pixels. In particular, we would like to have strong regularisation
across pixels that stem from events at approximately the same time, whereas
regularisation between pixels whose events have very different timestamps should
be reduced. This corresponds to a grouping of pixels in the time domain, based
on the timestamps of the recorded events. Solving computer vision problems on a
surface is also known as \emph{intrinsic image processing} \cite{Lai2011}, as it
involves the intrinsic (\ie coordinate-free) geometry  of the surface, a topic
studied by the field of differential geometry. Looking at the body of literature
on intrinsic image processing on surfaces, we can divide previous work into two
approaches based on the representation of the surface. Implicit approaches
\cite{Krueger2008,Cheng2002} use an implicit surface (e.g. through the zero
level set of a function), whereas explicit approaches \cite{Lui2008,Stam2003}
construct a triangular mesh representation. Our method uses the same underlying
theory of differential geometry, however we note that because the surface of
active events is defined by the timestamps which are monotonically increasing,
the class of surfaces is effectively restricted to $2\tfrac{1}{2}$D. This means
that there exists a simple parameterisation of the surface and we can perform
all computations in a local euclidean coordinate frame (\ie the image domain
$\Omega$). In contrast to \cite{Lai2011}, where the authors deal with arbitrary
surfaces, we avoid the need to explicitly construct a representation of the
surface. This has the advantage that we can straightforwardly make use of
GPU-accelerated algorithms to solve the large-scale optimisation problem. 
A similar approach was proposed recently in the context of variational
stereo \cite{Graber2015}.

We start by defining the surface $S\subset \mathbb{R}^3$ as the graph of a
scalar function $t(x,y)$ through the mapping $\varphi\ :\ \Omega \to S$
\begin{equation}
  \label{eq:1}
  X = \varphi(x,y) = \begin{bmatrix}x,&y,&t(x,y) \end{bmatrix}^T,
\end{equation}
where $X\in S$ denotes a 3D-point on the surface.
$t(x,y)$ is simply an image that records for each pixel $(x,y)$ the time since
the last event.
The partial derivatives
of the parameterisation $\varphi$ define a basis for the tangent space
$T_X\mathcal{M}$ at each point $X$ of the manifold $\mathcal{M}$, and the dot
product in this tangent space gives the \emph{metric} of the manifold. In
particular, the \emph{metric tensor} is defined as the symmetric $2\times 2$
matrix
\begin{equation}
  \label{eq:2}
  g=\begin{bmatrix} \langle \varphi_x, \varphi_x \rangle &&  \langle \varphi_x, \varphi_y
    \rangle \\
    \langle \varphi_x, \varphi_y \rangle &&  \langle \varphi_y, \varphi_y \rangle  \end{bmatrix},
\end{equation}
where subscripts denote partial derivatives 
and $\langle\cdot,\cdot\rangle$ denotes the scalar product. Starting from the
definition of the parameterisation \refEqn{eq:1}, straightforward calculation
gives $\varphi_x = \begin{bmatrix}1& 0& t_x\end{bmatrix}^T,\ \varphi_y =
\begin{bmatrix}0& 1& t_y\end{bmatrix}^T$ and
\begin{subequations}
  \begin{align}
    g&=\begin{bmatrix} 1+t_x^2 & t_xt_y \\ t_xt_y
    & 1+t_y^2 \end{bmatrix}\\
    g^{-1} &= \frac{1}{G}\begin{bmatrix} 1+t_y^2 & -t_xt_y \\ -t_xt_y
    & 1+t_x^2 \end{bmatrix}\ ,
  \end{align}
\end{subequations}
where $G=\operatorname{det}(g)$.

Given a smooth function $\tilde f\in C^1(S,\mathbb{R}) $ on the manifold, the
gradient of $\tilde f$ is characterised by $\mathrm{d}\tilde f(Y) = \langle
\nabla_{g} \tilde f, Y\rangle_g\quad \forall Y\in T_X\mathcal{M}$
\cite{Lee1997}. We will use the notation $\nabla_g \tilde f$ to emphasise the
fact that we take the gradient of a function defined on the surface (i.e. under
the metric of the manifold). $\nabla_g \tilde f$ can be expressed in local
coordinates as
\begin{equation}
  \label{eq:4}
  \nabla_g \tilde f = \left(g^{11}\tilde f_x + g^{12}\tilde f_y\right)\varphi_x + \left(g^{21}\tilde f_x + g^{22}\tilde f_y\right)\varphi_y,
\end{equation}
where $g^{ij},\ i,j=1,2$ denotes the components of the inverse of $g$ (the
so-called pull-back). Inserting $g^{-1}$
into \refEqn{eq:4} gives an expression
for the gradient of a function $\tilde f$ on the manifold in local coordinates
\begin{equation}\label{eqn:cont_nabla_g}
\nabla_g \tilde f = \frac{1}{G}\left[\left(\left(1+t_y^2\right)\tilde f_x - t_xt_y\tilde f_y\right)\begin{bmatrix}1& 0& t_x\end{bmatrix}^T +
  \left(\left(1+t_x^2\right)\tilde f_y - t_xt_y\tilde f_x\right)\begin{bmatrix}0& 1& t_y\end{bmatrix}^T\right]\ .
\end{equation}
Equipped with these definitions, we are ready to define
our regularisation term. It will be a variant of the total variation (TV) norm
insofar that we take the norm of the gradient of $\tilde f$ on the manifold
\begin{equation}
  \label{eq:regularisation_term}
  TV_g(\tilde f) = \int_S |\nabla_g \tilde f|\,\mathrm{d}s.
\end{equation}
It is easy to see that if we have $t(x,y)=const$, then $g$ is the $2\times 2$
identity matrix and $TV_g(\tilde f)$ reduces to the standard TV. Also note
that in the definition of the $TV_g$ we integrate over the surface.
Since our goal is to formulate everything in local coordinates, we relate
integration over $S$ and integration over $\Omega$ using the pull-back
\begin{equation}
  \label{eq:6}
  \int_S |\nabla_g \tilde f|\,\mathrm{d}s = \int_\Omega |\nabla_g \tilde f| \sqrt{G}\,\mathrm{d}x\mathrm{d}y,
\end{equation}
where $\sqrt{G}$ is the differential area element that links distortion of the
surface element $\mathrm{d}s$ to local coordinates $\mathrm{d}x\mathrm{d}y$. In
the same spirit, we can pull back the data term defined on the manifold to the
local coordinate domain $\Omega$. In contrast to the method of Graber
\etal~\cite{Graber2015} which uses the differential area element as
regularization term, we formulate the full variational model on the manifold,
thus incorporating spatial as well as temporal information.

\begin{figure}[t!]
  \begin{center}
    \subfigure[Flat surface]{\includegraphics[width=0.32\textwidth]{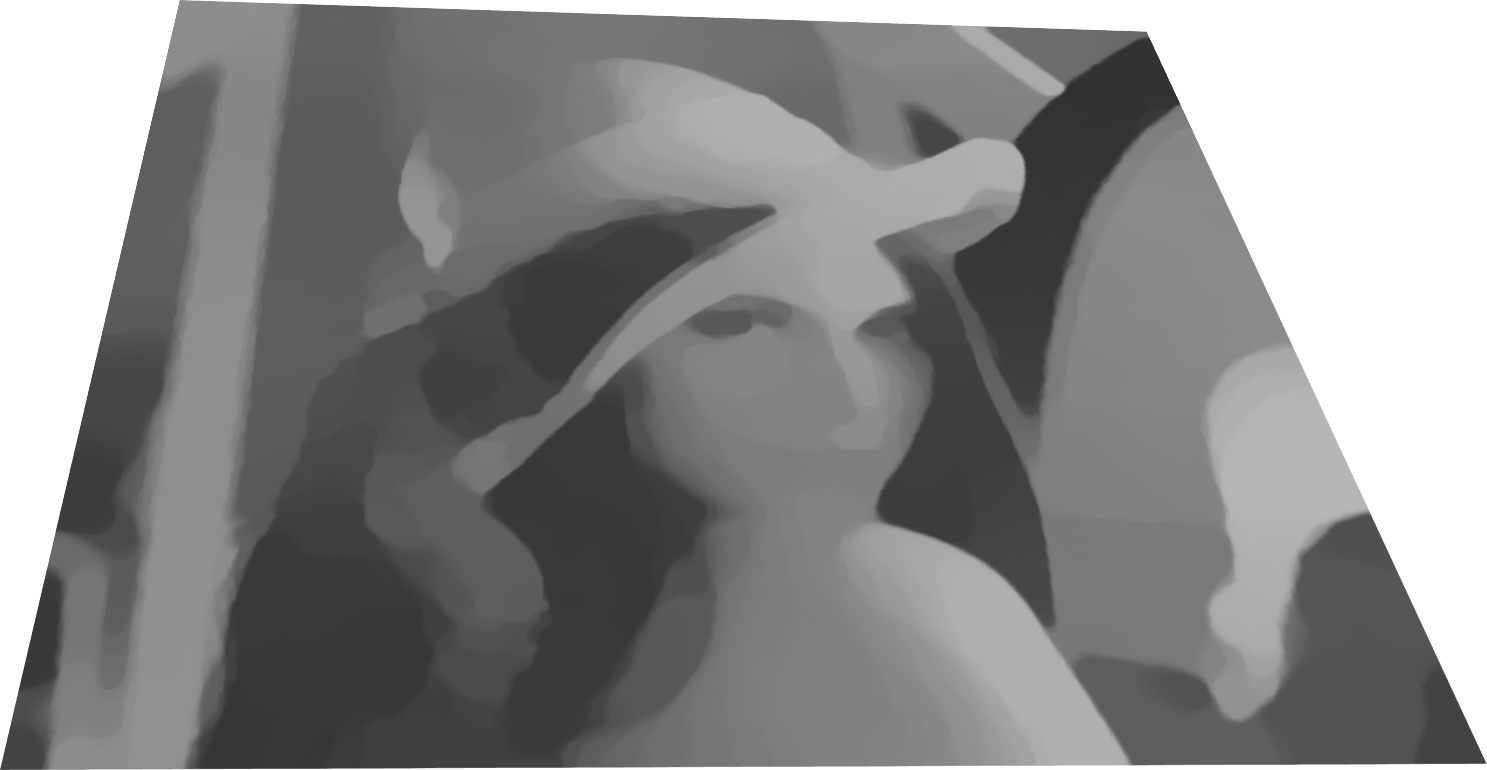} \label{subfig:flat}}
    \subfigure[Ramp surface]{\includegraphics[width=0.32\textwidth]{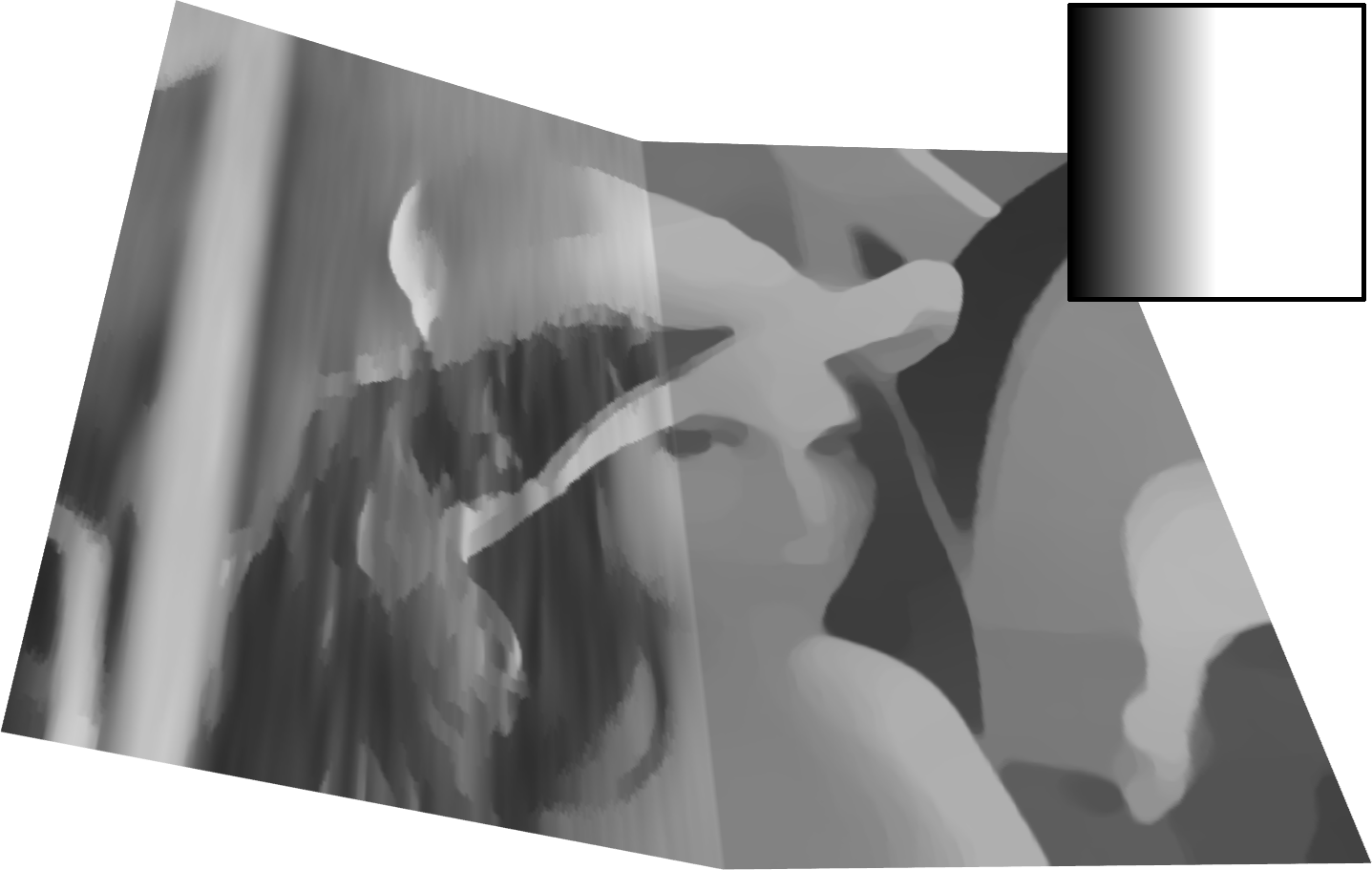}\label{subfig:ramp}}
    \subfigure[Sine surface]{\includegraphics[width=0.32\textwidth]{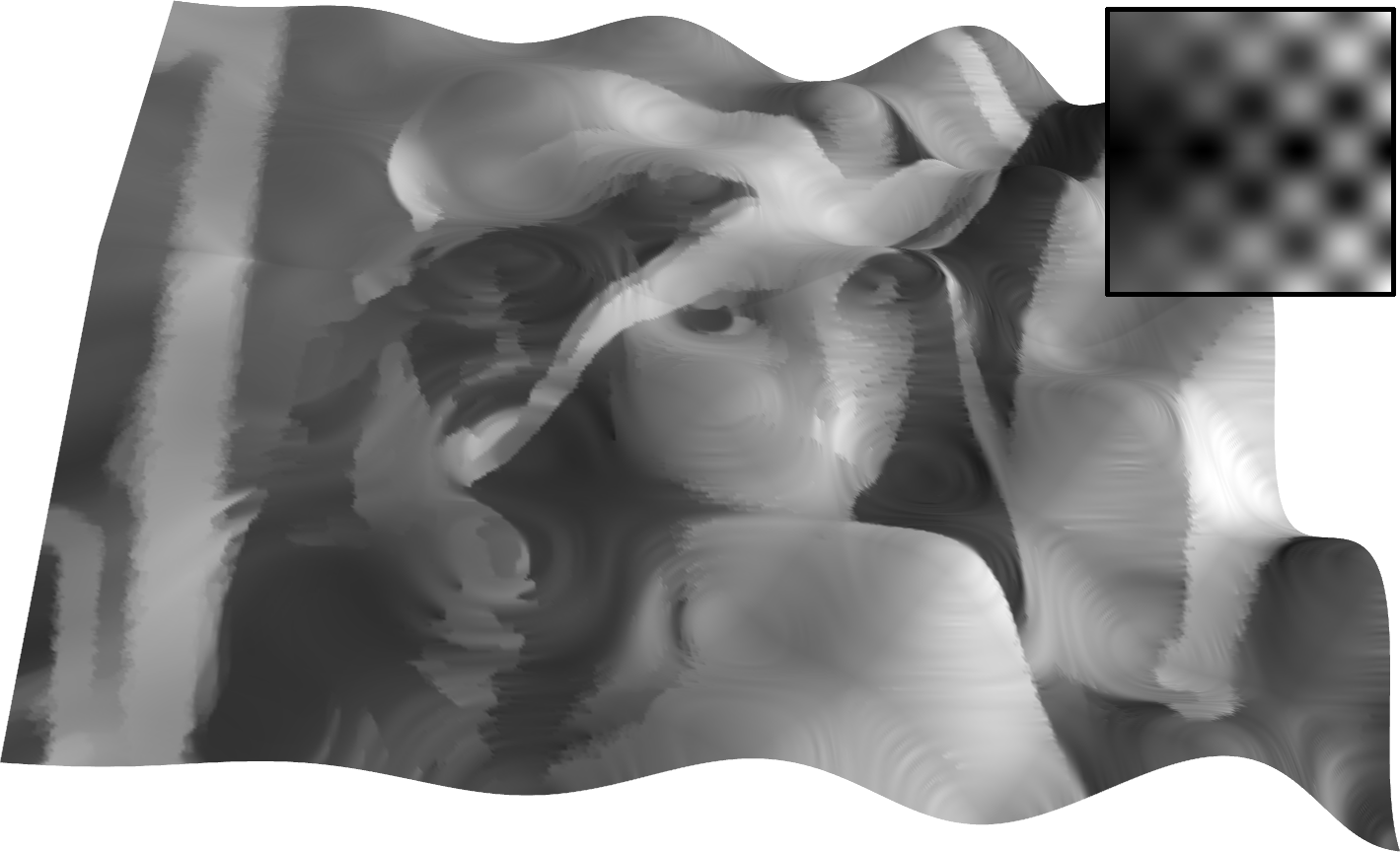}\label{subfig:sine}}
    \caption{ROF denoising on different manifolds. A flat surface \subref{subfig:flat} gives
      the same result as standard ROF denoising, but more complicated surfaces
      \subref{subfig:ramp}\subref{subfig:sine} significantly change
      the result. The graph function $t(x,y)$ is depicted in the upper right corner. We can see that
      a ramp surface \subref{subfig:ramp} produces regularisation anisotropy due to the fact that the
      surface gradient is zero in $y$-direction but non-zero in
      $x$-direction. The same is true for the sine  surface \subref{subfig:sine}, where we can see
      strong regularisation along level sets of the surface and less regularisation across level sets.}
    \label{fig:rof_manifold}
  \end{center}
\end{figure}

To assess the effect of $TV_g$ as a regularisation term, we depict in
\refFig{fig:rof_manifold} results of the following variant of the ROF denoising
model \cite{Rudin1992}
\begin{equation}
  \label{eq:7}
  \min_u \int_\Omega |\nabla_g u|\sqrt{G} + \tfrac{\lambda}{2}|u-f |^2\sqrt{G}\mathrm{d}x\mathrm{d}y,
\end{equation}
with different $t(x,y)$, \ie ROF-denoising on different manifolds.
We see that computing the TV norm on the manifold can be interpreted as
introducing anisotropy based on the surface geometry (see Fig.
\ref{subfig:ramp},\ref{subfig:sine}). We will use this to guide regularisation
of the reconstructed image according to the surface defined by the event time.

\paragraph*{Data Term}
The data term $D(u,f^n)$ encodes the deviation of $u$ from the noisy measurement
$f^n$ \refEqn{eqn:intensity_update}. Under the reasonable assumption that a
neuromorphic camera sensor suffers from the same noise as a conventional sensor,
the measured update caused by one event will contain noise. In computer vision,
a widespread approach is to model image noise as zero-mean additive Gaussian.
While this simple model is sufficient for many applications, real sensor noise
is dependent on scene brightness and should be modelled as a Poisson
distribution \cite{Ratner2007}. We therefore define our data term as
\begin{equation} \label{eqn:data_term}
   D(u,f^n) :=\lambda \int_{S} \left(u-f^n\log u\right)\mathrm{d}s = \lambda \int_\Omega \left(u - f^n\log u\right)\sqrt{G}\,\mathrm{d}x\mathrm{d}y\quad
    \text{s.t. } u(x,y) \in [u_\mathrm{min},u_\mathrm{max}]
\end{equation}
whose minimiser is known to be the correct ML-estimate under the assumption of
Poisson-distributed noise between $u$ and $f^n$ \cite{Le2007}. Note that, in
contrast to \cite{Graber2015}, we also define the data term to lie on the
manifold. \refEqn{eqn:data_term} is also known as {\em generalised
Kullback-Leibler divergence} and has been investigated by Steidl and Teuber
\cite{Steidl2010} in variational image restoration methods. Furthermore, the
data term is convex, which makes it easy to incorporate into our variational
energy minimisation framework. We restrict the range of $u(x,y) \in
[u_\mathrm{min},u_\mathrm{max}]$ since our reconstruction problem is defined up
to a gray value offset caused by the unknown initial image intensities.

\paragraph*{Discrete Energy}
In the discrete setting, we represent images of size $M\times M$ as matrices in
$\mathbb{R}^{M\times M}$ with indices $(i,j) = 1\ldots M$. Derivatives are
represented as linear maps $L_x,L_y\ :\ \mathbb{R}^{M\times M}\to
\mathbb{R}^{M\times M}$, which are simple first order finite difference
approximations of the derivative in $x$- and $y$-direction \cite{Chambolle2004}.
The discrete version of $\nabla_g$, defined in \refEqn{eqn:cont_nabla_g}, can
then be represented as a linear map $L_g\ :\ \mathbb{R}^{M\times
M}\to\mathbb{R}^{M\times M \times 3}$ that acts on $u$ as follows
  \begin{subequations}
    \label{eq:3}
    \begin{align}
      (L_gu)_{ij1} &= \tfrac{1}{G_{ij}}\left((1+(L_yt)_{ij}^2)(L_xu)_{ij} - (L_xt)_{ij}(L_yt)_{ij}(L_yu)_{ij}\right)\nonumber \\
      (L_gu)_{ij2} &=\tfrac{1}{G_{ij}}\left((1+(L_xt)_{ij}^2)(L_yu)_{ij} - (L_xt)_{ij}(L_yt)_{ij}(L_xu)_{ij}\right)\nonumber \\
      (L_gu)_{ij3} &= \tfrac{1}{G_{ij}}\left( (L_xt)_{ij}(L_xu)_{ij} + (L_yt)_{ij}(L_yu)_{ij} \right)\nonumber 
    \end{align}
  \end{subequations}
  Here, $G\in\mathbb{R}^{M\times M}$ is the pixel-wise determinant of $g$ given
  by $G_{ij}=1+(L_xt)_{ij}^2+(L_yt)_{ij}^2$. The discrete data term follows
  from \refEqn{eqn:data_term} as $D(u,f^n):=\lambda
  \sum_{i,j}(u_{ij}-f^n_{ij}\log u_{ij})\sqrt{G_{ij}}$. This yields the complete
  discrete energy
\begin{equation}
  \label{eq:5}
  \min_u \|L_gu\|_{g}+\lambda \sum_{i,j}\left(u_{ij}-f^n_{ij}\log u_{ij}\right)\sqrt{G_{ij}}\quad
  \text{s.t. } u_{ij} \in [u_\mathrm{min},u_\mathrm{max}],
\end{equation}
with the $g$-tensor norm defined as $\|A\|_{g} = \sum_{i,j}\sqrt{G_{ij}\sum_l
(A_{ijl})^2}\quad \forall A\in\mathbb{R}^{M\times M\times 3}$.

\subsection{Minimising the Energy}
\label{sec:minimization}

We minimise \eqref{eq:5} using the Primal-Dual algorithm \cite{Chambolle2011}.
Dualising the $g$-tensor norm yields the primal-dual formulation
\begin{equation}\label{eqn:primal_dual_energy}
  \min_u \max_p \big[D(u,f^n) + \langle L_gu,p\rangle -R^*(p) \big],
\end{equation}
where $u\in \mathbb{R}^{M\times M}$ is the discrete image,
$p\in\mathbb{R}^{M\times M\times 3}$ is the dual variable and $R^*$ denotes the
convex conjugate of the $g$-tensor norm. A solution of
\refEqn{eqn:primal_dual_energy} is obtained by iterating
\begin{subequations}\label{eqn:cont_iterates}
\begin{align}
  u_{k+1} =& (I+\tau \partial{D})^{-1}(u_k-\tau L_g^*p_k) \nonumber\\
  p_{k+1} =& (I+\sigma \partial {R}^*)^{-1}(p_k+\sigma L_g(2 u_{k+1} - u_k)), \nonumber
\end{align}
\end{subequations}
where $L_g^*$ denotes the adjoint operator of $L_g$. The proximal maps for the
data term and the regularisation term can be solved in closed form, leading to
the following update rules
\begin{subequations}\label{eqn:update_rules}
\begin{align}
  \hat u&=\operatorname{prox}_{\tau D}(\bar{u}) &\Leftrightarrow \hat u_{ij}&=
  \underset{u_\mathrm{min},u_\mathrm{max}}{\operatorname{clamp}}\left(
  \tfrac{1}{2}\left(\bar{u}_{ij}-\beta_{ij}+
         \sqrt{\left(\bar{u}_{ij}-\beta_{ij}\right)^2+4\beta_{ij}
             f^n_{ij}}\right)\right) \nonumber \\
  \hat p &= \operatorname{prox}_{\sigma R^*}(\bar{p}) &\Leftrightarrow \hat p_{ijl} &=
   \frac{\bar{p}_{ijl}}{\max\lbrace1,\nicefrac{\|\bar{p}_{ij,\cdot}\|}{\sqrt{G_{ij}}}\rbrace}, \nonumber
\end{align}
\end{subequations}
with $\beta_{ij} = \tau\lambda\sqrt{G_{ij}}$. The time-steps $\tau,\sigma$ are
set according to $\tau\sigma \leq \nicefrac{1}{\|L_g\|^2}$, where we estimate
the operator norm as $\|L_g\|^2\leq 8+4\sqrt{2}$. Since the updates are
pixel-wise independent, the algorithm can be efficiently parallelised on GPUs.
Moreover, due to the low number of events added in each step, the algorithm
usually converges in $k\leq 50$ iterations.


\section{Experiments}
We perform our experiments using a DVS128 camera with a spatial resolution of
$128 \times 128$ and a temporal resolution of \unit[1]{$\mu s$}. The parameter
$\lambda$ is kept fixed for all experiments. The thresholds
$\Delta^+,\Delta^-$ are set according to the chosen camera settings. In
practice, the timestamps of the recorded events can not be used directly as the
manifold defined in \refSec{sec:manifold} due to noise. We therefore denoise the
timestamps with a few iterations of a TV-L1 denoising method. We compare our
method to the recently proposed method of \cite{Bardow2016} on sequences
provided by the authors. Furthermore, we will show the influence of the proposed
regularisation on the event manifold using a few self-recorded sequences.

\subsection{Timing}
In this work we aim for a real-time reconstruction method. We implemented the
proposed method in C++ and used a Linux computer with a \unit[3.4]{GHz}
processor and a NVidia Titan X GPU\footnote{We note that the small image size of
$128 \times 128$ is not enough to fully load the GPU such that we measured
almost the same wall clock time on a NVidia 780 GTX Ti.}. Using this setup we
measure a wall clock time of \unit[\textbf{1.7}]{\textbf{ms}} to create one
single image, which amounts to \unit[$\approx 580$]{fps}. While we can create a
new image for each new event, this would create a tremendous amount of images
due to the number of events ($\approx 500.000$ per second on natural scenes with
moderate camera movement). Furthermore one is limited by the monitor refresh
rate of \unit[60]{Hz} to actually display the images. In order to achieve
real-time performance, one has two parameters: the number of events that are
integrated into one image and the number of frames skipped for display on
screen. The results in the following sections have been achieved by accumulating
$500$ events to produce one image, which amounts to a time resolution of
\unit[3-5]{ms}.
\subsection{Influence of the Event Manifold}
\begin{figure}[t!]
\begin{center}
	\includegraphics[width=0.18\textwidth]{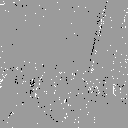}
	\includegraphics[width=0.18\textwidth]{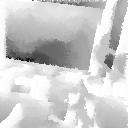}
	\includegraphics[width=0.18\textwidth]{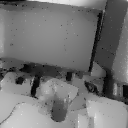}
	\includegraphics[width=0.18\textwidth]{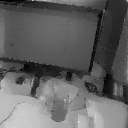}\\
	\includegraphics[width=0.18\textwidth]{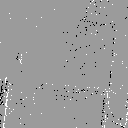}
	\includegraphics[width=0.18\textwidth]{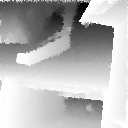}
	\includegraphics[width=0.18\textwidth]{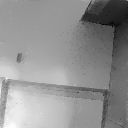}
	\includegraphics[width=0.18\textwidth]{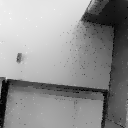}\\
	\caption{Sample results from our method. The columns depict raw events, time
manifold, result without manifold regularisation and finally with our manifold
regularisation. Notice the increased contrast in weakly textured regions
(especially around the edge of the monitor).}
	\label{fig:manifold}
\end{center}
\end{figure}

We have captured a few sequences around our office with a DVS128 camera. In
\refFig{fig:manifold} we show a few reconstructed images as well as the raw
input events and the time manifold. For comparison, we switched off the manifold
regularisation (by setting $t(x,y)=const$), which results in images with notably
less contrast.

\subsection{Comparison to Related Methods}
In this section we compare our reconstruction method to the method proposed by
Bardow \etal~\cite{Bardow2016}. The authors kindly provided us with the recorded
raw events, as well as intensity image reconstructions at regular timestamps
$\delta t=15 \mathrm{ms}$. Since we process shorter event packets, we search for
the nearest neighbour timestamp for each image of \cite{Bardow2016} in our
sequences. We visually compare our method on the sequences {\em face}, {\em
jumping jack} and {\em ball} to the results of \cite{Bardow2016}. We point out
that no ground truth data is available so we are limited to purely qualitative
comparisons.

\begin{figure}[ht]
\begin{center}
  \includegraphics[width=0.15\textwidth]{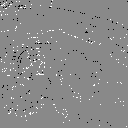}
	\includegraphics[width=0.15\textwidth]{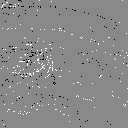}
	\includegraphics[width=0.15\textwidth]{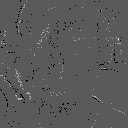}
	\includegraphics[width=0.15\textwidth]{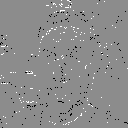}
	\includegraphics[width=0.15\textwidth]{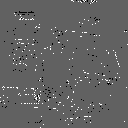}
	\includegraphics[width=0.15\textwidth]{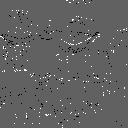}\\
	\includegraphics[width=0.15\textwidth]{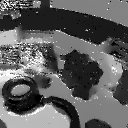}
	\includegraphics[width=0.15\textwidth]{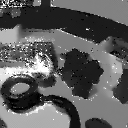}
	\includegraphics[width=0.15\textwidth]{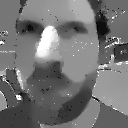}
	\includegraphics[width=0.15\textwidth]{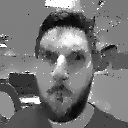}
	\includegraphics[width=0.15\textwidth]{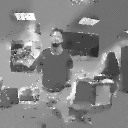}
	\includegraphics[width=0.15\textwidth]{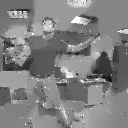}\\
  \includegraphics[width=0.15\textwidth]{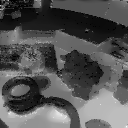}
	\includegraphics[width=0.15\textwidth]{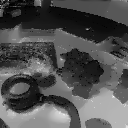}
	\includegraphics[width=0.15\textwidth]{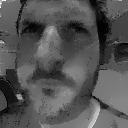}
	\includegraphics[width=0.15\textwidth]{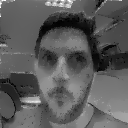}
	\includegraphics[width=0.15\textwidth]{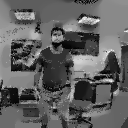}
	\includegraphics[width=0.15\textwidth]{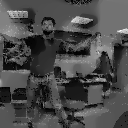}
	\caption{Comparison to the method of \cite{Bardow2016}. The first row shows
	the raw input events that have been used for both methods. The second row
	depicts the results of Bardow \etal, and the last row shows our result. We can
	see that out method produces more details (e.g. face, beard) as well as more
	graceful gray value variations in untextured areas, where \cite{Bardow2016}
	tends to produce a single gray value.}
	\label{fig:comparison_bardow}
\end{center}
\end{figure}

In \refFig{fig:comparison_bardow} we show a few images from the sequences. Since
we are dealing with highly dynamic data, we point the reader to the included
supplementary video\footnote{\url{https://www.youtube.com/watch?v=rvB2URrGT94}}
which shows whole sequences of several hundred frames.

\subsection{Comparison to Standard Cameras}
\begin{figure}[h!]
\begin{center}
	\includegraphics[width=0.22\textwidth]{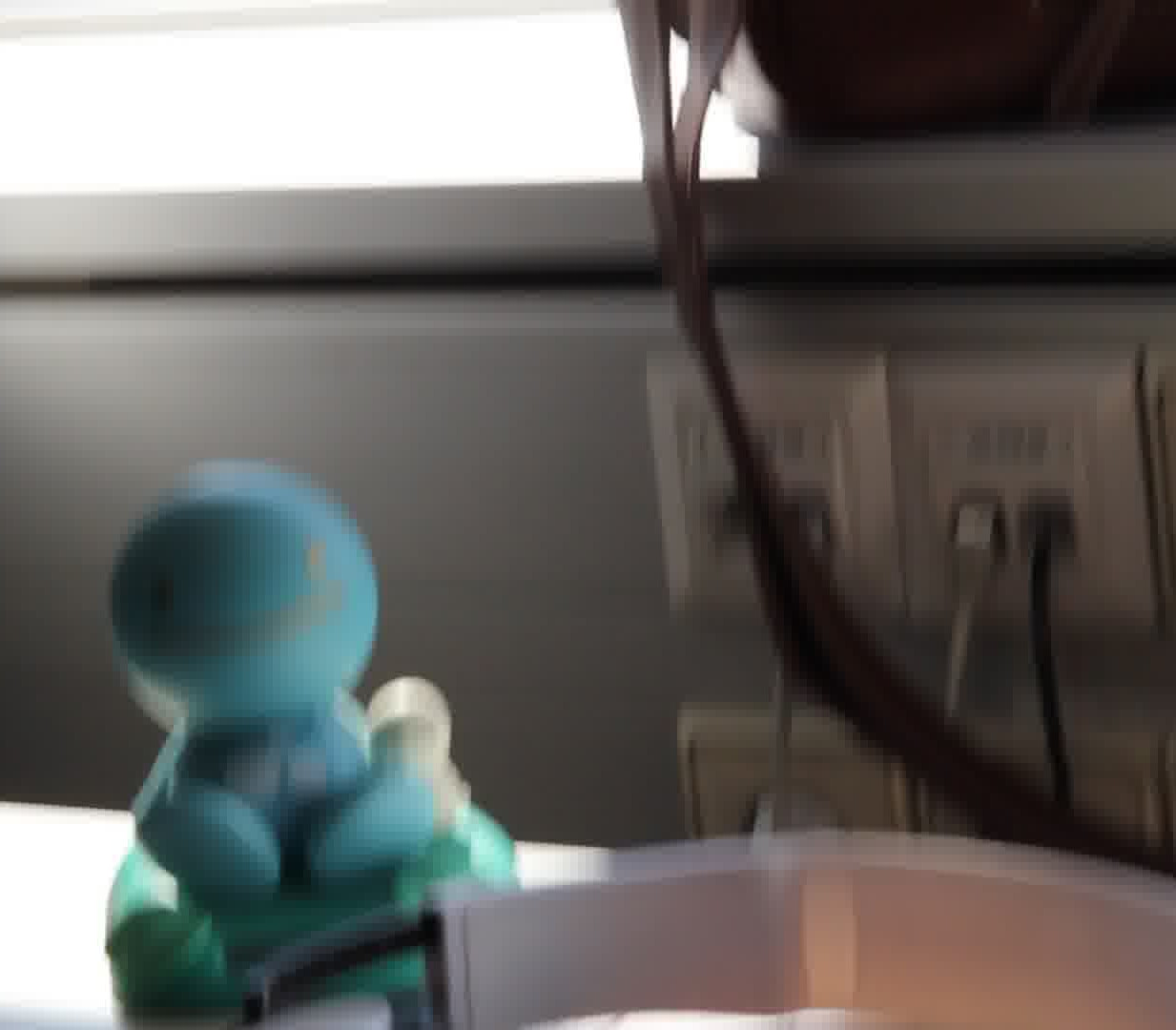}
	\includegraphics[width=0.22\textwidth]{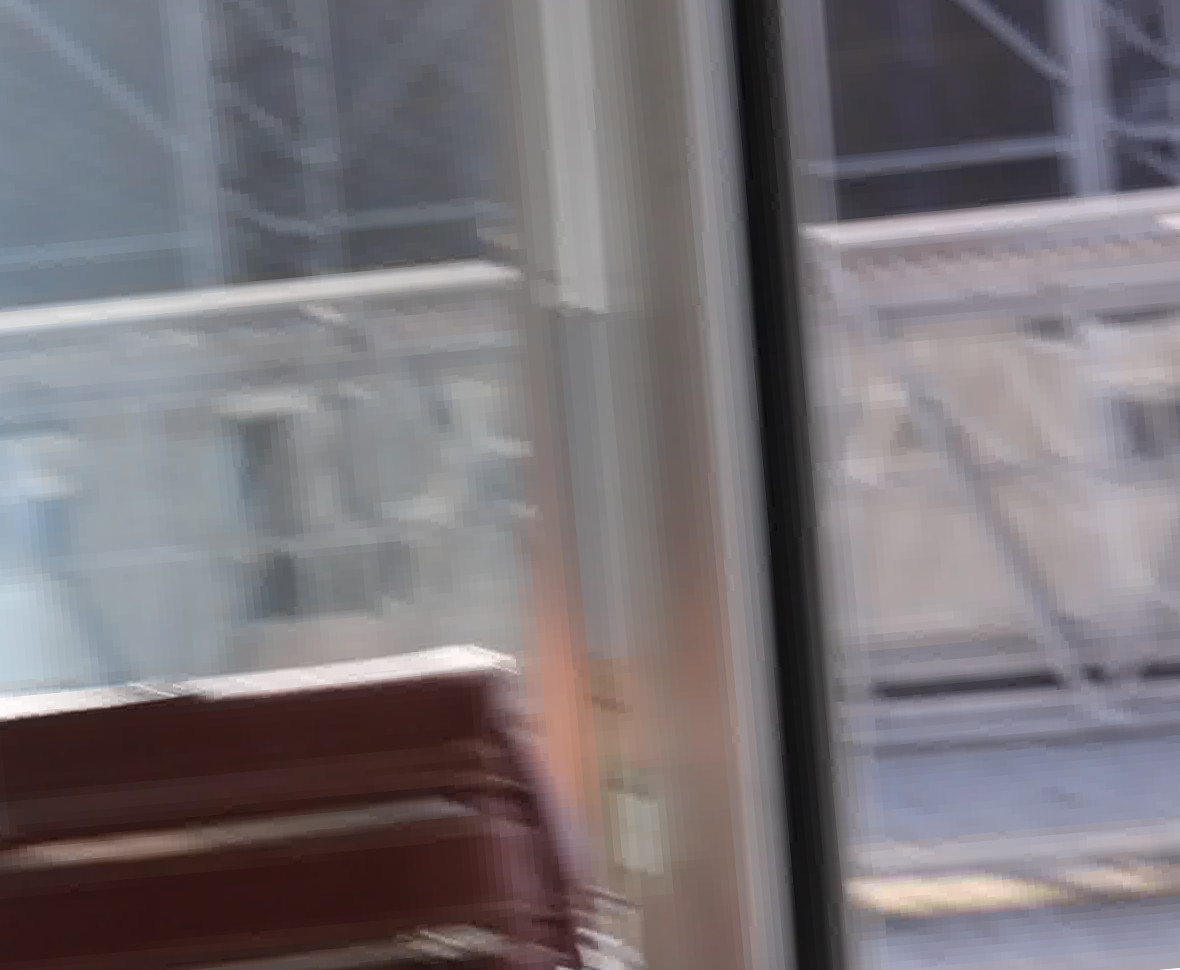}
	\includegraphics[width=0.22\textwidth]{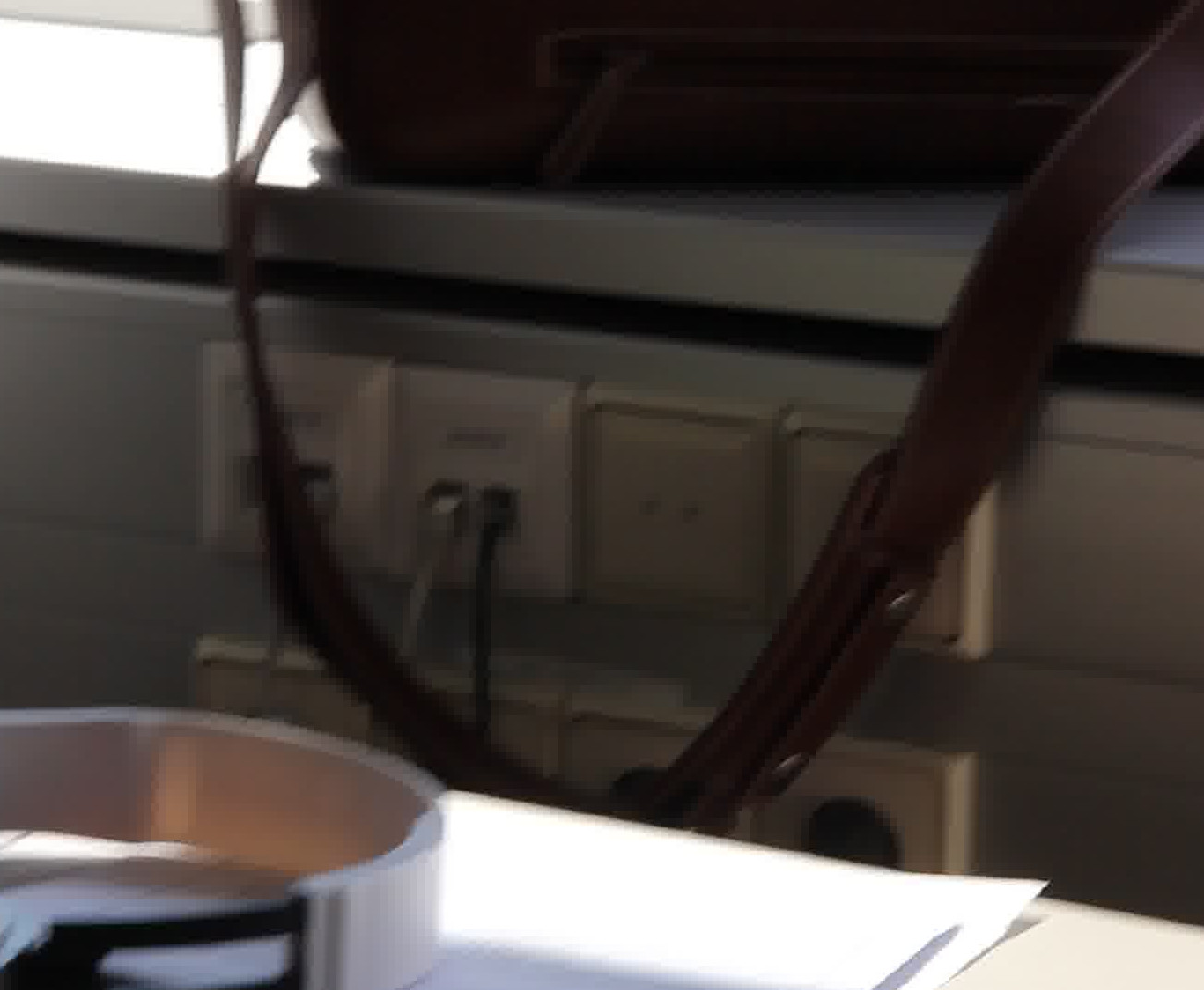}
	\includegraphics[width=0.22\textwidth]{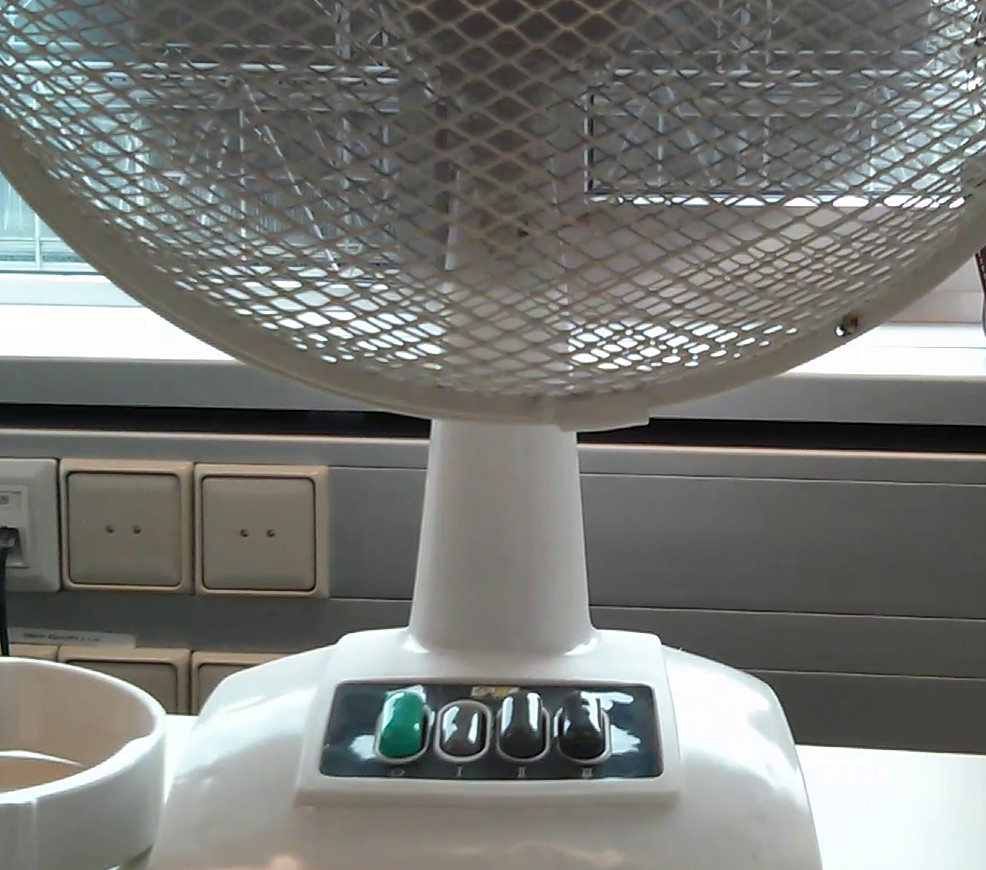}\\
	\includegraphics[width=0.22\textwidth]{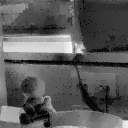}
	\includegraphics[width=0.22\textwidth]{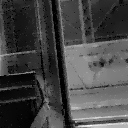}
	\includegraphics[width=0.22\textwidth]{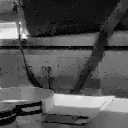}
	\includegraphics[width=0.22\textwidth]{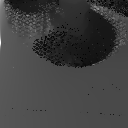}
	\caption{Comparison to a video captured with a modern DSLR camera. Notice the
	rather strong motion blur in the images of the DSLR (top row), whereas the DVS
	camera can easily deal with fast camera or object movement (bottom row).}
	\label{fig:comparison_camera}
\end{center}
\end{figure}

We have captured a sequence using a DVS128 camera as well as a Canon EOS60D DSLR
camera to compare the fundamental differences of traditional cameras and
event-based cameras. As already pointed out by \cite{Bardow2016}, rapid movement
results in motion blur for conventional cameras, while event-based cameras show
no such effects. Also the dynamic range of a DVS is much higher, which is also
shown in \refFig{fig:comparison_camera}.

\section{Conclusion}
In this paper we have proposed a method to recover intensity images from
neuromorphic or event cameras in real-time. We cast this problem as an iterative
filtering of incoming events in a variational denoising framework. We propose to
utilise a manifold that is induced by the timestamps of the events to guide the
image restoration process. This allows us to incorporate information about the
relative ordering of incoming pixel information without explicitly estimating
optical flow like in previous works. This in turn enables an efficient algorithm
that can run in real-time on currently available PCs.

Future work will include the study of the proper noise characteristic of event
cameras. While the current model produces natural-looking intensity images, a
few noisy pixels appear that indicate a still non-optimal treatment of sensor
noise within our framework. Also it might be beneficial to look into a local
minimisation of the energy on the manifold (\eg by coordinate-descent) to
further increase the processing speed.

\section*{Acknowledgements}
This work was supported by the research initiative Mobile Vision with funding
from the AIT and the Austrian Federal Ministry of Science, Research and Economy
HRSM programme (BGBl. II Nr. 292/2012).

\bibliography{egbib}
\end{document}